\PassOptionsToPackage{table}{xcolor}
\documentclass[10pt, a4paper]{article}

\usepackage[final]{lrec2026} % this is the new style
\usepackage{algorithm}
\usepackage{algorithmic}
\usepackage{booktabs}
\usepackage{tikz}
\usepackage{tikz-dependency}
\usepackage{tabularx}
\usepackage{ragged2e}
\usepackage{multirow}
\usepackage{xcolor}
\usepackage{fancyvrb}
\usepackage{framed}
\usepackage[ragged]{sidecap}
\usepackage{lineno}
\usepackage{hyperref}
\usepackage{url}
\usepackage{amsfonts}       % blackboard math symbols
\usepackage{nicefrac}       % compact symbols for 1/2, etc.
\usepackage{microtype}      % microtypography
\usepackage{xspace}
\usepackage{amsmath}
\usepackage{amssymb}
\usepackage{graphicx}
\usepackage{multicol}
\usepackage[noabbrev,capitalize,nameinlink]{cleveref}
\usepackage[most]{tcolorbox}
\usepackage{wrapfig}
\usepackage{subcaption}
\usepackage{newfloat}
\usepackage{listings}

\floatstyle{ruled}
\newfloat{listing}{tb}{lst}{}
\floatname{listing}{Listing}

\newcommand*\circled[1]{\tikz[baseline=(char.base)]{
            \node[shape=circle,draw,inner sep=.6pt] (char) {#1};}}

\tcbset{
    promptstyle/.style={
        enhanced,
        width=\linewidth,
        colback=white,
        colframe=black,
        colbacktitle=gray!20,
        coltitle=black,
        rounded corners,
        boxrule=0.5pt,
        drop shadow=black!50!white,
        attach boxed title to top left={
            xshift=-2mm,
            yshift=-2mm
        },
        boxed title style={
            rounded corners,
            size=small,
            colback=gray!20
        }
    },
    replystyleg/.style={
        enhanced,
        width=\linewidth,
        colback=green!15,
        colframe=black,
        colbacktitle=green!30,
        coltitle=black,
        boxrule=0.5pt,
        drop shadow=black!50!white,
        rounded corners,
        sharp corners=north,
        attach boxed title to top right={
            xshift=-2mm,
            yshift=-2mm
        },
        boxed title style={
            rounded corners,
            size=small,
            colback=green!40
        }
    },
    replystyler/.style={
        enhanced,
        width=\linewidth,
        colback=red!15,
        colframe=black,
        colbacktitle=red!40,
        coltitle=black,
        boxrule=0.5pt,
        drop shadow=black!50!white,
        rounded corners,
        sharp corners=north,
        attach boxed title to top right={
            xshift=-2mm,
            yshift=-2mm
        },
        boxed title style={
            rounded corners,
            size=small,
            colback=red!40
        }
        }
    }

\newtcolorbox{promptbox}[1][]{
    promptstyle,
    title=Prompt,
    #1
}

\newcommand{\sefl}{\textsc{SEFL}}
\newcommand{\llamabig}{\texttt{Llama-3.1-70B}}
\newcommand{\qwenbig}{\texttt{Qwen2.5-72B}}
\newcommand{\qwenmini}{\texttt{Qwen2.5-0.5B}}
\newcommand{\llamamini}{\texttt{Llama-3.2-1B}}
\newcommand{\llamasmall}{\texttt{Llama-3.2-3B}}
\newcommand{\llamamed}{\texttt{Llama-3.1-8B}}
\newcommand{\qwenmed}{\texttt{Qwen2.5-14B}}

\title{\sefl{}: A Framework for Generating Synthetic Educational Assignment Feedback with LLM Agents}

\name{Mike Zhang\textsuperscript{{\rm $\ddagger$}{\rm $\dagger$}{\rm $\diamond$}}, Amalie Pernille Dilling\textsuperscript{\rm $\dagger$}, Léon Gondelman\textsuperscript{\rm $\dagger$}\\ {\bf \large Niels Erik Ruan Lyngdorf\textsuperscript{\rm $\dagger$}, Euan D. Lindsay\textsuperscript{\rm $\dagger$}, Johannes Bjerva\textsuperscript{\rm $\dagger$}}}

\address{\textsuperscript{\rm $\ddagger$}University of Copenhagen, Denmark \\
         \textsuperscript{\rm $\dagger$}Aalborg University, Denmark \\
         \textsuperscript{$\diamond$}Pioneer Centre for AI, Denmark \\
         {\tt mike.zhang@di.ku.dk}\\}

\abstract{
Providing high-quality feedback on student assignments is crucial for student success, but it is heavily limited by time and budgetary constraints. 
In this work, we introduce \textbf{S}ynthetic \textbf{E}ducational \textbf{F}eedback \textbf{L}oops (SEFL), a synthetic data framework designed to generate data that resembles immediate, on-demand feedback at scale without relying on extensive, real-world student assignments and teacher feedback. 
To obtain this type of data, two large language models (LLMs) operate in a teacher-student role to simulate assignment completion and formative feedback, generating 19.8K synthetic pairs of student work and corresponding critiques and actionable improvements from a teacher. 
With this data, we fine-tune smaller, more computationally efficient LLMs on these synthetic pairs, enabling them to replicate key features of high-quality, goal-oriented feedback. 
Through comprehensive evaluations with three LLM judges and three human experts, across a subset of 900 outputs, we demonstrate that SEFL-tuned models outperform both their untuned counterparts and an existing baseline in terms of feedback quality. 
The potential for societal impact is reinforced by extensive qualitative comments and ratings from human stakeholders — both students and higher education instructors. SEFL has the potential to transform feedback processes for higher education and beyond.
 \\ \newline }

\begin{document}

\maketitleabstract

\section{Introduction}
Constructive feedback is a cornerstone of higher education, promoting critical thinking and fostering deeper understanding~\cite{hattie2008visible, costello2013technologies}. 
In higher education settings, however, providing consistent, high-quality feedback is complicated by privacy, consent, and transparency considerations in data collection~\cite{fischer2020mining, suresh-etal-2022-talkmoves, demszky-hill-2023-ncte, wang-demszky-2024-edu, wang2024tutor,lindsay2024responsibledevelopmentautomatedstudent}, in addition to being a labor-intensive task. 
Advances in language technology offer opportunities to automate and augment higher education feedback processes, addressing these limitations.

In particular, LLMs have shown progress in education~\cite{wang2024large}, including automated grading~\cite{ke2019automated, ramesh2022automated, stahl-etal-2024-exploring} and personalized tutoring~\cite{yun2024enhancing, liu2024personality, rooein2024conversations, ross-andreas-2024-toward, kwon2024biped, 19b02384b88f4404b0a2f5d5eec1207f, 2babc566a8034738ac303e9ba612df14,wang-etal-2024-book2dial}. Yet, automating teacher-student assignment feedback with LLMs %in agentic, dialogic settings~\cite{xi2023rise, guo2024large, zhang2024simulating}
remains an open question.
We seek to answer: \textbf{RQ.} \emph{How can synthetic teacher-student interactions generated by LLMs be leveraged to enable scalable and effective educational student assessment feedback?}

\begin{figure}
    \centering
    \includegraphics[width=\linewidth]{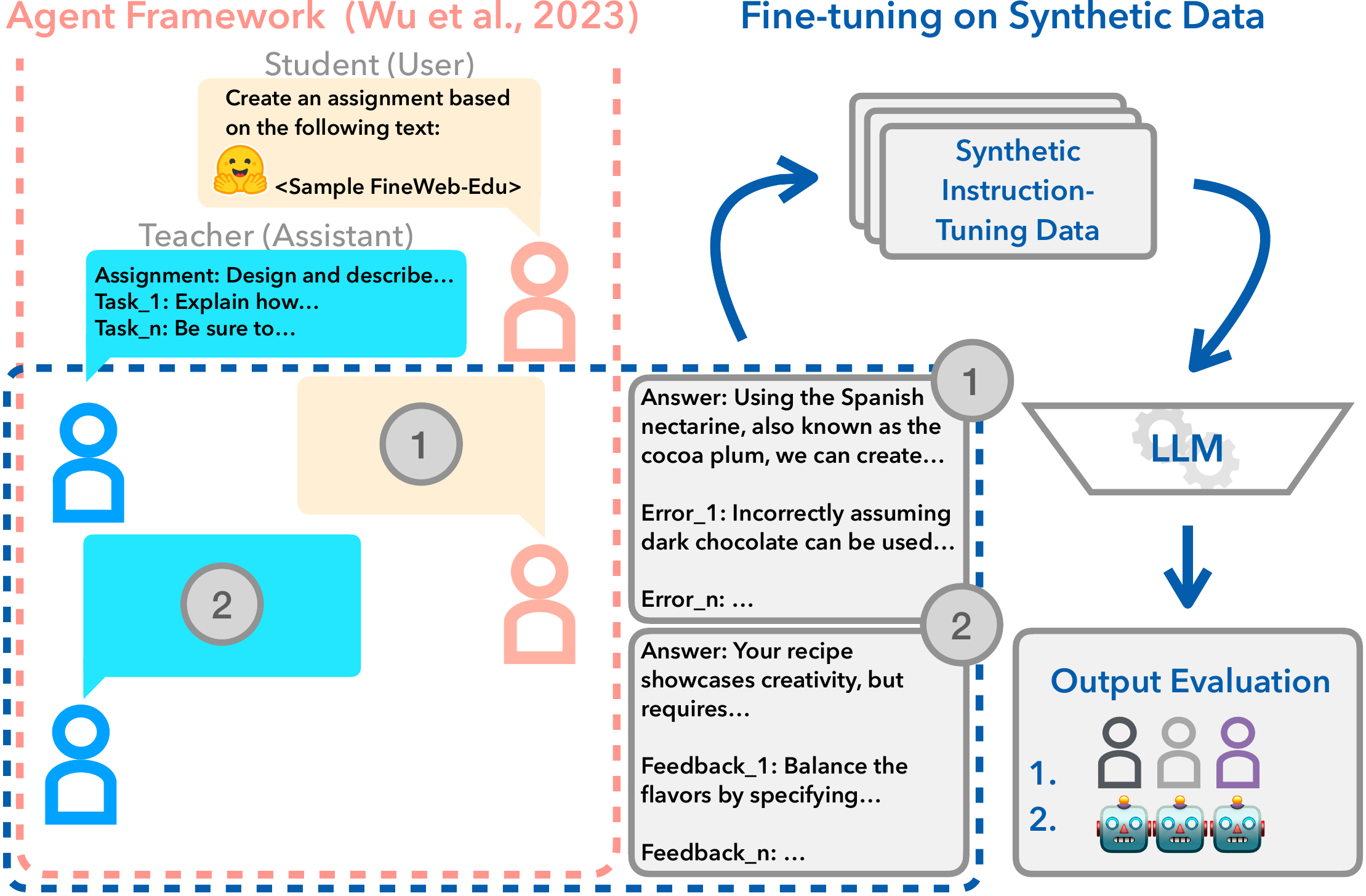}
    \caption{\justifying\textbf{\sefl{} Synthetic Data Generation Setup.} We use a two-agent framework~\cite{wu2023autogen} with LLMs acting as a Student and Teacher. The Teacher creates assignments from Fineweb-Edu~\cite{lozhkov2024fineweb-edu}, a dataset curated using LLMs to judge the educational value of web pages. Overall, the Student responds with explicit errors (via prompting), and finally, the Teacher addresses each mistake. This synthetic interaction data is then used to fine-tune multiple LLMs, whose performance is measured through human ratings and evaluations by LLMs-as-judges.
    }
    \label{fig:fig1}
\end{figure}

Here, we introduce \textbf{S}ynthetic \textbf{E}ducational \textbf{F}eedback \textbf{L}oops (\sefl{}), a framework that generates synthetic teacher-student interactions using LLM agents. 
In this framework, two LLMs, one acting as the teacher and the other as the student, simulate \emph{formative} feedback workflows~\cite{conole2006contemporary, nicol2007assessment}.
This synthetic data from the agents is then used to fine-tune smaller autoregressive language models, resulting in better feedback models that enable the development of scalable educational feedback systems. These systems can operate efficiently on the modest computational infrastructure available in higher education institutions, without requiring access to privacy-sensitive data. 

We show that:
(i) Empirically, SEFL-tuned models outperform their non-tuned versions in win rate evaluations by three LLM judges and three human experts in giving assessment feedback. 
(ii) Through empirical and qualitative analysis of expert annotator comments, we find that larger models tend to provide more actionable, goal-oriented, and user-friendly feedback. They are also more consistent and better at supporting student autonomy.
(iii) We observe strong agreement among human evaluators regarding feedback quality. 
(iv) We compare our approach to an existing baseline (Book2Dial;~\citealp{wang-etal-2024-book2dial}) and show that SEFL-tuned models provide better feedback according to all LLM judges.

\paragraph{Contributions.} We contribute the following:
\begin{itemize}
\itemsep0em
\item \textbf{SEFL}: An agentic framework that simulates teacher-student feedback loops with paired language model agents.  
\item 19,841 synthetic assignment--feedback pairs generated by SEFL to fine-tune smaller language models.
\item A comprehensive set of human and LLM evaluations that outline the strengths and limitations of SEFL with extensive qualitative analysis of the feedback provided by human experts.
\item Open-sourcing of all models, code, and data.
\footnote{Resources and Supplementary Material such as full prompts and example outputs can be found at \url{https://github.com/jjzha/sefl} and \url{https://tinyurl.com/3zdu847k} (HuggingFace).}  
\end{itemize}

\section{Related Work}
\paragraph{NLP \& Education.} 
Large language models are now supporting a broad spectrum of educational tasks. 
In automated grading, they score short answers, essays, and even programming assignments with accuracy that approaches expert instructors, which eases faculty workload and releases time for mentoring \cite{ke2019automated,ramesh2022automated,stahl-etal-2024-exploring}. 
For personalized tutoring, conversational agents powered by these models adapt explanations, hints, and examples to each learner’s background knowledge and preferred style, producing measurable gains in engagement and achievement \cite{yun2024enhancing,liu2024personality,rooein2024conversations,ross-andreas-2024-toward,kwon2024biped,19b02384b88f4404b0a2f5d5eec1207f}. 
Research on peer learning shows that the same technology can mediate small-group discussions, suggest prompts, and highlight diverse viewpoints, leading to richer collaboration \cite{bauer2023using}. In mathematics, aligning word problems and proofs to grade-level objectives has been automated, yielding encouraging results and reducing the manual effort required to curate question banks \cite{botelho2023leveraging}. Critical thinking curricula also benefit; LLMs can challenge students to justify claims, detect fallacies, and refine arguments in real-time \cite{guerraoui-etal-2023-teach}. The models have also begun to assist scholars: studies report successful use for screening literature, summarizing drafts, and aligning reviewer comments with revision plans \cite{liang2024can,sonkar2024pedagogical}. Complementing these functions are analytics tools that track learning trajectories and surface early warnings when a student slips behind \cite{schwarz2018orchestrating,aslan2019investigating,alrajhi2021urgency}.

Despite this growing body of work, prior studies have rarely targeted the systematic generation of feedback on open-ended student submissions and assignments. In this study, we address this gap by using LLMs to generate extensive comment sets that teachers can accept as is or adapt. Decades of scholarship define effective feedback as goal-oriented, actionable, timely, user-friendly, and consistent while fostering self-evaluation \cite{carless2011developing,wiggins2012seven}. Feedback that is long and verbose can confuse learners. Therefore, concise wording is usually preferable, and comments that arrive soon after the original effort drive steady improvement \cite{wiggins2012seven}. By producing immediate responses and tailoring suggestions to rubric criteria, LLM-based systems stand well-positioned to satisfy these guidelines while operating at the classroom and institution scale.

\paragraph{Synthetic Data Frameworks.} Recent research shows how collaborative agentic LLMs can synthesize large-scale interactional datasets for educational tasks. For example, CAMEL~\cite{li2023camel} employs cooperative role-based dialogues to achieve shared objectives, whereas SimSeek~\cite{kim-etal-2022-generating} utilizes agent-based conversations to construct comprehensive information-seeking datasets. In education, SocraticLM~\cite{liu2024socraticlm} simulates Socratic tutoring through multi-turn dialogue, and Book2Dial~\cite{wang-etal-2024-book2dial} generates teacher-student conversations from textbooks. \emph{In contrast, SEFL focuses on concise teacher-student feedback loops rather than extended instructional dialogues}. While~\citet{nair-etal-2024-closing} explore iterative revisions, SEFL generates diverse feedback pairs from assignment-answer-feedback tuples, enabling fine-tuning of smaller, cost-effective models for large-scale use.

\section{SEFL: Synthetic Educational Feedback Loops}

\subsection{Synthetic Data Generation}
We employ a two-agent framework~\cite{wu2023autogen} to simulate a student-teacher feedback loop, as seen in higher education. 
Both the teacher and student roles are simulated by two separate \llamabig{} models for a two-turn conversation.\footnote{Note that if we mention a model, it is always the \textit{post-trained} version (i.e., \texttt{-Instruct}).} The models are tasked to generate assignment$\rightarrow$answer$\rightarrow$feedback tuples. 
First, the student-agent requests an assignment using Fineweb-Edu~\cite{lozhkov2024fineweb-edu} texts (\cref{fig:fig1}), which is known for its educational content based on LLM judgments. 
Second, the teacher-agent creates an assignment that can be of any domain, e.g., STEM, social sciences, puzzles, and so forth. 
Then, the student-agent (\circled{1}) submits assignments containing several explicit errors, and the teacher-agent (\circled{2}) provides feedback. In \circled{3} and \circled{4}, we obtain an assignment-feedback pair. 
We deliberately avoided imposing strict rules on the types of errors that could be generated. We hypothesize that allowing the LLM to introduce errors naturally results in a more diverse and realistic set of mistakes than if we had a fixed checklist. 
We investigate both \qwenbig{} and \llamabig{} for interactions. 
For a control check, we generate 5,000 interaction tuples with each model, and validate the output as an initial step to investigate the quality of initial feedback.

We show in~\cref{tab:valid} the results of this experiment. 
Out of 5,000 generated examples, \llamabig{} generates 2,513 valid examples (i.e., valid \texttt{JSON} format and each feedback point refers to an error) compared to \qwenbig{} with 454 valid examples. 
For a further check, we use BERTScore~\cite{Zhang2020BERTScore} as a proxy to see whether each error-feedback pair of the valid generations relates to each other.\footnote{We only calculate it of the samples where both error and feedback have the same number of generations.} 
We show that, regardless of \llamabig{} generating more valid examples, the BERTScore (0.877) remains in a similar range to \qwenbig{} (0.919); in both cases, this indicates a high level of similarity. 
As a final qualitative check, we experimented with several prompts and consolidated them into a single, final prompt, which is included in the supplementary material. 
Finally, we use \llamabig{}-generated data as the basis for all subsequent model fine-tuning, as it generated more valid examples. We spot-checked several prompts and consolidated the final full prompt in~\cref{fig:prompt} (\cref{sec:prompts}).

\begin{table}[t]
    \centering
    \footnotesize
    \begin{tabular}{lrr}
    \toprule
                    & Valid (/ 5,000) & BERTScore     \\
                    \midrule
    \llamabig{}     & 2,513             & 0.877         \\
    \qwenbig{}      & 454               & 0.919         \\
    \bottomrule
    \end{tabular}
    \caption{\textbf{Generation Capabilities.} First, We show the number of valid examples, measured by correct \texttt{JSON} format and whether each feedback refers to an error. \llamabig{} generates more valid examples. Second, we measure BERTScore as a proxy for relatedness between error--feedback pairs of the valid generations.}
    \label{tab:valid}
\end{table}

\paragraph{Data statistics.} 
After generating 5,000 examples, we continue to create example pairs, resulting in 19,841 teacher-student feedback pairs. 
In \cref{tab:stats}, we present the final dataset statistics. 
We underline that the generation lengths for each agent are intentionally kept concise ($<$170 subword tokens), based on the hypothesis that overly lengthy feedback may be counterproductive.
This is in line with observations from \citet{ferguson2011student}, who finds that students tend to favor brief comments. 
We argue that balancing supportive and critical feedback is crucial, as LLMs often produce verbose responses by default, which can influence the preferences of both humans and LMs~\cite{saito2023verbosity}.

\paragraph{Task Errors.} 
Once all data was generated, we took a random subsample of 226 instances with 428 explicit errors. We investigated what category of \emph{intentional} errors the student agent generates.
In total, we identified seven error types from the 428 errors and show the proportion in~\cref{tab:full_error_summary}. Most errors are of type ``Omission and/or incompleteness'' (surmounted for 50\% of the errors):
\begin{enumerate}
    \itemsep0em
    \item \textbf{Omission and/or incompleteness:} This includes missing details, lacking examples or evidence, not exploring a topic deeply enough, or failing to address all parts of the assignment.
    \item \textbf{Stylistic and/or formatting issues:} Problems with grammar, spelling, punctuation, tone (e.g., too informal), word count, and incorrect formatting (like improper citations).
    \item \textbf{Factual inaccuracy:} This includes incorrect dates, numbers, names, scientific facts, historical events, or misquoting a source text.
    \item \textbf{Logical flaws and/or weak argumentation:} These are errors in reasoning. They include making incorrect assumptions, oversimplifying complex topics, drawing faulty conclusions, or failing to construct a coherent and well-supported argument.
    \item \textbf{Technical and/or procedural errors:} This category is for mistakes in technical execution, such as incorrect mathematical calculations, flawed experimental design, or errors in code.
    \item \textbf{Structural and organizational problems:} These errors relate to the overall structure and flow of the response, such as a lack of clear transitions between paragraphs, poor organization, or a missing introduction/conclusion.
    \item \textbf{Task misinterpretation:} Student agent misunderstands the core requirement of the task, such as writing a story instead of an essay or failing to use a requested framework.
\end{enumerate}

\begin{table*}[t]
\centering
\scriptsize
\begin{tabularx}{\textwidth}{p{4cm} r r X}
\toprule
\textbf{Error Category} & \textbf{Count} & \textbf{\%} & \textbf{Example Error Explanation from Student Agent} \\
\midrule
Omission and/or Incompleteness & 214 & 50.0\% & ``Failure to provide specific examples and evidence to support claims''. \\\midrule
Stylistic and/or Formatting Issues & 67 & 15.7\% & ``Incorrect spelling of the word 'protection' in the last sentence of the answer''. \\\midrule
Factual Inaccuracy & 59 & 13.8\% & ``Incorrectly claimed Ibn al-Haytham was born in Egypt instead of Basra, Iraq''. \\\midrule
Logical Flaws and/or Weak Argumentation & 53 & 12.4\% & ``The statement about DSP chips having limitations is an oversimplification, as they are still widely used for various applications''. \\\midrule
Technical and/or Procedural Errors & 16 & 3.7\% & ``Incorrect calculation of the target waiting time, should be a 30\% reduction from 10 minutes''. \\\midrule
Structural and/or Organizational Problems & 12 & 2.8\% & ``Lack of clear structure and organization in the comparative analysis''. \\\midrule
Task Misinterpretation & 7 & 1.6\% & ``The writing deploys facts that were not mentioned in the letter, like United States' interest in containing the spread of communism''. \\
\midrule
\textbf{Total} & \textbf{428} & \textbf{100.0\% }& --- \\
\bottomrule
\end{tabularx}
\caption{\textbf{Statistics of Explicit Error Types.} We show the explicit error types. We show that omission and incompleteness are proportionally the most used \emph{intentional} errors the student agent generates.}
\label{tab:full_error_summary}
\end{table*}

\section{Methodology}

\subsection{Fine-Tuning Large Language Models}
We divide the data into 17,856 training examples and 1,985 validation examples. 
To test the feedback quality of our approach across model scales, we fine-tune five open-weight models, namely \qwenmini{}, \llamamini{}, \llamasmall{}, \llamamed{}, and \qwenmed{} on this synthetic dataset. The compute we train the models on is AMD Radeon Instinct MI250X GPUs, and it took a total of 467 GPU hours.

\paragraph{Training Objective.}
To fine-tune the LLMs, for each prompt $x$ and target sequence $y=(y_1,..,y_T)$ we minimize the token-level cross-entropy

\begin{equation}
\mathcal{L}_{\text{SFT}}(\theta)= -\sum_{t=1}^{T} m_t \,\log p_{\theta} \bigl(y_t \mid y_{<t},x\bigr),
\end{equation}

where $m_t$ masks out the prompt tokens and activates the loss on reference tokens only.

\paragraph{Optimization Details.}
All models train for three epochs with a global batch size of 16 and context lengths of 131K for Qwen2.5 and 128K for the Llama variants. 
We use AdamW with $\beta_1=0.9$, $\beta_2=0.999$~\cite{loshchilov2018decoupled}, $\epsilon=10^{-8}$, weight decay \(0.1\), and gradient clipping at norm \(1.0\). 
The learning rate peaks at \(2\times10^{-5}\) after a linear warm-up covering the first five percent of steps and then follows a linear decay.

\begin{table}[t]
    \centering
    \small
    \begin{tabular}{lr}
    \toprule
    \textbf{Feature}                             & \textbf{Value}                     \\
    \midrule
    Instances                                    & 19,841                             \\
    Assignment Length                            & 78.6                               \\
    Length (Student Agent)                       & 168.1                              \\
    \hspace{1.5em} \# Errors Points              & 2.5                                \\
    \hspace{1.5em} Length \# Errors              & 20.7                               \\
    Length (Teacher Agent)                       & 120.5                              \\
    \hspace{1.5em} \# Feedback Points            & 2.5                                \\
    \hspace{1.5em} Length \# Feedback            & 34.6                               \\
    \bottomrule
    \end{tabular}
    \caption{\textbf{Generation Statistics.} We show the dataset statistics in \emph{averages}, where length is measured in whitespace-separated tokens.}
    \label{tab:stats}
\end{table}

\subsection{Multi-faceted Evaluation}\label{subsec:eval}
\subsubsection{Human Evaluation}
After fine-tuning the language models with the assignment-feedback pairs and testing the performance of \sefl{}, we conduct a human evaluation on a total of 450 instances with three annotators, following a similar approach to SoctraticLM~\cite{liu2024socraticlm}, which annotated around 1,000 dialogues with 10 annotators.
We randomly sample the subset of instances from the validation set. 
For each item, both the original instruction tuned model (A) and the model further fine-tuned with \sefl{} (B) produce feedback. Three human experts compared pairs of feedback responses produced for the same assignment and answer. For each item they read the original prompt, the student submission, and the two candidate feedback drafts from the non-tuned and \sefl-tuned model. Then, they select the feedback from model A or B that is better based on four base criteria:

\begin{itemize}
    \itemsep0em
    \item \textbf{Accuracy:} The generated feedback text focuses on concrete strengths and weaknesses in the student's work, avoiding superficial remarks.
    \item \textbf{Actionability:} Suggestions are clear, specific, and realistic for a student to apply.
    \item \textbf{Conciseness:} Wording is brief and focused, with little repetition.
    \item \textbf{Tone:} Language stays constructive and professional while recognizing good elements.
\end{itemize}

Raters were reminded to value efficiency over length, to prefer targeted advice over general principles, and to ignore formatting tricks.  
They recorded their choice as \texttt{A} or \texttt{B} and could leave an optional free-text comment. With this, we calculate the win rate (i.e., the percentage of choosing one feedback text over the other). This has become a de facto standard to evaluate long-form text against each other (e.g., \citealp{rafailov2023direct}).

Each row took at most ten minutes, and the guidelines stressed taking regular breaks to sustain attention. We deliberately remove the $A=B$ tie option because a forced choice provides more informative labels and reduces hesitation. At the same time, a separate checkbox still allows raters to mark assignment $\rightarrow$ answer $ \rightarrow$ feedback tuples as unrelated or nonsensical. This happened around 12\% of the time, especially in the smaller, less capable models. The full annotation guidelines are reported in the supplementary material (\cref{tab:annotation_guidelines}; \cref{app:humaneval}).

\paragraph{Annotator Demographics.} Our raters are aged 20-40 and from Europe. One identifies as female and the other two identify as male. One female and one male have a background in Computer Science, and one male has a background in Engineering Education. All have extensive experience in teaching and supervision, or being taught and supervised; they all work in higher education (at various levels, such as research assistant and assistant professors) with near-native English proficiency.

\begin{figure}
    \centering
    \includegraphics[width=\linewidth]{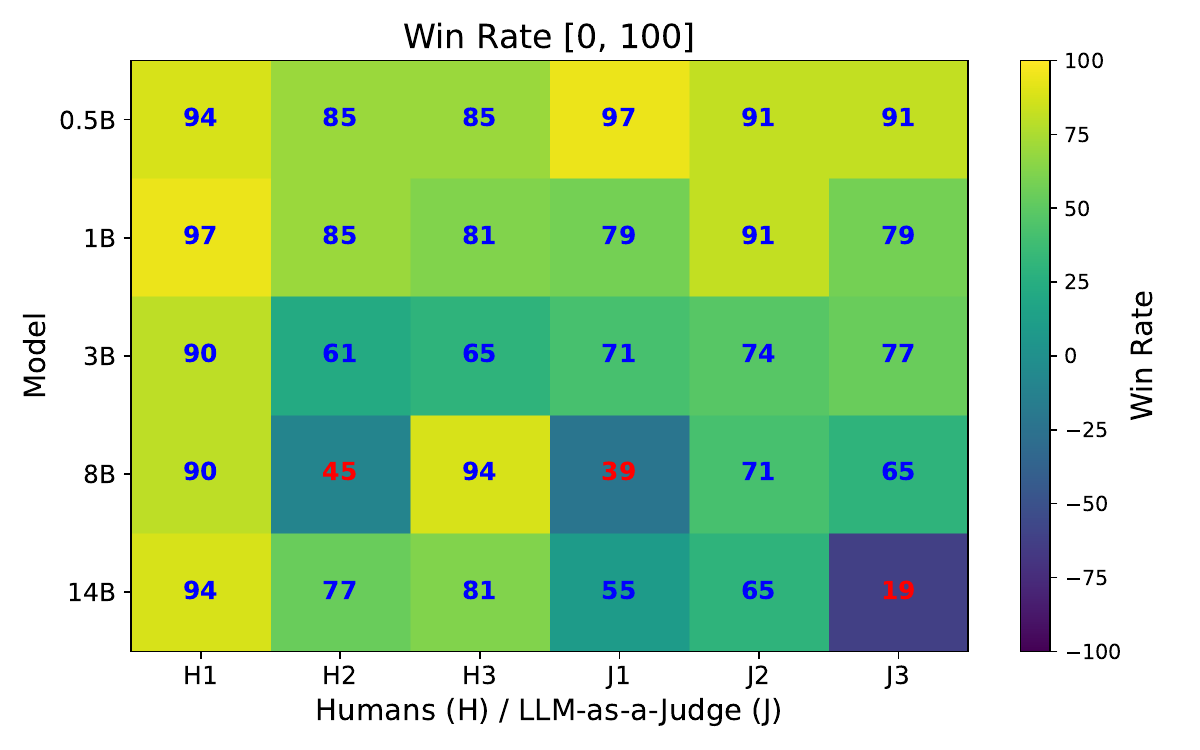}
    \caption{\textbf{Win Rate Results.} We show the win rate of our \emph{SEFL-tuned models}. A win rate $>$50\% indicates that \sefl-tuned models are better in giving feedback than their vanilla counterpart; in red everything $<$50\% shows the opposite. We show results of 3 human annotators (H\#) and 4 LLM judges: \texttt{gpt-4o} (J1), \texttt{claude-3.5-sonnet} (J2), and \texttt{deepseek-v3} (J3).}
    \label{fig:results}
\end{figure}

\subsubsection{LLM-as-a-Judge} 
We also evaluate the fine-tuned models' output using a LLM-as-a-judge framework, a method gaining traction for evaluating free text output~\cite{liu-etal-2023-g,zheng2024judging,chen-etal-2023-exploring-use,verga2024replacing,tornberg2023chatgpt,naismith-etal-2023-automated,gilardi2023chatgpt,kocmi-federmann-2023-large,huang-etal-2024-chatgpt, gu2024surveyllmasajudge,falkandchen2025how}. 
The three LLMs rate the same 450 random instances, namely GPT-4o~\cite{hurst2024gpt}, Claude3.5-Sonnet, and DeepSeek-V3~\cite{liu2024deepseek}. For the LLM-as-a-judge experiments using closed-source models, we used the respective APIs, and the total cost was approximately 10 USD.

For every example, the judge model receives the assignment prompt together with the two candidate feedback drafts.  
It is asked to decide which draft is better on the same four base criteria as the human annotators.  
The instruction forbids numeric grades or explanations and requires the judge to output exactly one character, \texttt{A} or \texttt{B}, producing a clean pairwise preference label. %We report the full prompt in the supplementary material in the code repository.

\begin{figure}[t]
    \centering
    \includegraphics[width=\linewidth]{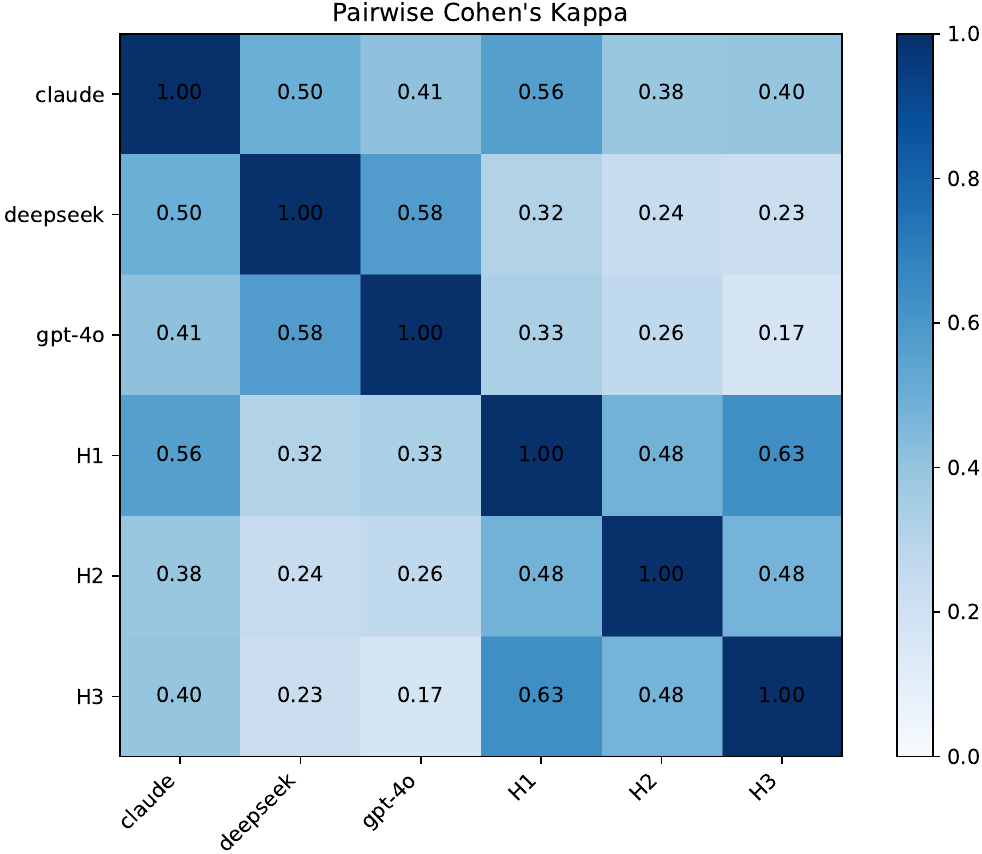}
    \caption{\textbf{Pairwise Cohen's $k$.} We show the pairwise Cohen's $k$ between each LLM judge and annotator.}
    \label{fig:kappa}
\end{figure}

\begin{figure*}[t]
\centering
\begin{tcolorbox}[enhanced,
  width=\linewidth,
  colback=white,
  boxrule=0.6pt,
  title=Qualitative feedback comparison: SEFL-tuned vs.\ Base (Qwen2.5-0.5B)]
\scriptsize

\begin{promptbox}
\textbf{Assignment (abridged).}
Summarize the lasting social, environmental, and economic fallout of the Bhopal disaster.
Explain (i) health effects on survivors, (ii) ongoing economic losses and compensation, and (iii) groundwater contamination and its ecological reach.

\textbf{Student answer (abridged).}
Mentions immediate deaths and chronic illness, links health to income loss, notes stigma in the local workforce, points out toxic wells but supplies no data, and omits policy or clean-up discussion.
\end{promptbox}

\begin{minipage}{0.53\linewidth}
\begin{tcolorbox}[promptstyle,title=SEFL-tuned feedback]
\begin{enumerate}
\item Recognises how the answer connects the three impact areas.
\item Requests figures on cancer incidence and water toxicity.
\item Links polluted wells to widening wealth gaps and gender effects.
\item Suggests mobile clinics and community water monitoring.
\item Encourages reflection on data-driven safety advocacy.
\end{enumerate}
\end{tcolorbox}
\end{minipage}\hfill
\begin{minipage}{0.45\linewidth}
\begin{tcolorbox}[promptstyle,title=Base model feedback]
\begin{enumerate}
\item Lists illnesses but echoes wording from the prompt.
\item Calls for better regulation yet cites no agencies.
\item Adds an unrelated section on industrial automation.
\item Repeats points in the conclusion and shifts tone.
\item Offers no actionable advice for revision.
\end{enumerate}
\end{tcolorbox}
\end{minipage}

\end{tcolorbox}
\caption{\textbf{Qualitative Example of Feedback.} Excerpt that shows how SEFL improves specificity and actionability. Full conversation will be added as supplementary material.}
\label{fig:qualitative-main}
\end{figure*}

\section{Results}
In \cref{fig:results}, we show the \emph{win rates} of models fine-tuned with \sefl{} vs.\ their non-tuned version, evaluated by both humans and LLM-based judges. A value above 50\% indicates that the \sefl-tuned models are preferred over their original versions. We show an example of the feedback in~\cref{fig:qualitative-main}, where we depict the abridged prompt and feedback by a tuned and non-tuned model.%(\cref{app:example}).

\paragraph{Human Assessment.} 
Overall, human rater evaluations in~\cref{fig:results} (H\#) show that the \sefl-tuned models often achieve high win rates, surpassing 90\% in several cases compared to the smaller models. The human annotators differed in their views on the 8B model's output quality; however, they generally converged on the observation that the fine-tuned 14B model produces superior feedback compared to its original version. By contrast, models not fine-tuned with \sefl{} had lower win rates, suggesting that \sefl{} provides an edge in generating more coherent and context-relevant feedback. In addition, we asked annotators whether the synthetic assignment$\rightarrow$answer$\rightarrow$feedback sequences were consistent. In over 75\% of cases, they confirmed the alignment between assignment, student response, and the feedback given, showing positive contextual relevance.

\paragraph{LLM-as-a-Judge Evaluation.} 
For the LLM-as-a-judge evaluations (J\#) in the same figure, we observe several differences in win rates depending on the model and scale. The results largely mirror the human assessment trend up to the 3B scale. The results from the three LLM judges (J1: GPT-4o, J2: Claude-3.5-Sonnet, J3: Deepseek-v3) reveal that \sefl-tuned models show varying levels of performance relative to their vanilla counterparts.\footnote{Models are picked based on their performance on RewardBench~\cite{lambert2024rewardbench}, JudgeBench~\cite{tan2024judgebench}, and JudgeArena~\cite{judge-arena}.} For instance, \qwenmini{} achieved the highest win rates across all three judges, indicating a consistent preference for the fine-tuned version. As the parameter size increases, the difference in win rate decreases (e.g., 58\% on average for 8B). The full judge prompt will be released in the supplementary material.

\paragraph{Human and Model Agreements.}
In~\cref{fig:kappa}, we present the pairwise Cohen's $k$ values~\cite{cohen1960coefficient} computed between each LLM judge and the human rater, to observe whether humans and LLM judges agree on which model gives better feedback.
The agreement among humans was moderate to substantial: H1 and H3 reached $\kappa = 0.63$, H1 and H2 $0.48$, and H2 and H3 $0.48$~\cite{landis1977measurement}.
Among the models, Claude aligns most closely with both the other judges and the humans; Deepseek follows, and GPT-4o shows the weakest match.
Across all model and human pairs, the numbers range from 0.17 to 0.58, which shows the subjectivity of feedback evaluation.
This suggests that human experts generally agree on the quality of feedback, but for LLM judges, this may vary. Overall, we see an opportunity for improvements in terms of agreement between LLMs and humans regarding feedback quality.

\begin{table*}[t]
\centering
\scriptsize

\rowcolors{2}{gray!20}{white}

\begin{tabularx}{\textwidth}{p{1.5cm} p{4.2cm} p{4.3cm} p{4.3cm}}
\toprule
\rowcolor{white}
\textbf{Model} & \textbf{H1 sample} & \textbf{H2 sample} & \textbf{H3 sample} \\
\midrule
\texttt{0.5B} &
``The feedback repeats the assignment rather than pointing out issues.'' &
``A is concise but misses important detail.'' &
``Comments do not reflect missing parts of the answer.'' \\

\texttt{0.5B-SEFL} &
``Feedback spots deliberate errors and offers clear fixes.'' &
``A is concise and clear, suggestions align with the rubric.'' &
``Advice notes that the discuss part is missing and proposes adding it.'' \\
\midrule
\texttt{1B} &
``Feedback echoes what the answer already states.'' &
``More precise.'' &
``Comments ignore omitted sections.'' \\

\texttt{1B-SEFL} &
``The note opens with strengths and gives concrete next steps.'' &
``Hints link directly to rubric points.'' &
``Advice adds two practical examples to guide revision.'' \\
\midrule
\texttt{3B} &
``Both models are good, A sounds nicer.'' &
``Both are good here.'' &
``Feedback covers content but wording feels harsh.'' \\

\texttt{3B-SEFL} &
``Feedback keeps an encouraging tone while giving actionable points.'' &
``Structure and tone feel balanced.'' &
``Comments cover content, style, and structure in one cohesive note.'' \\
\midrule
\texttt{7B} &
``Feedback addresses surface errors but misses deeper reasoning.'' &
``Detailed but drifts off the prompt.'' &
``Overlooks the rubric item about evidence.'' \\

\texttt{7B-SEFL} &
``Points out reasoning gaps and proposes specific fixes.'' &
``Stays focused on the prompt and remains concise.'' &
``Directly cites the missing evidence section.'' \\
\midrule
\texttt{14B} &
``Feedback is thorough but wording is opaque.'' &
``Long sentences reduce clarity.'' &
``Some comments repeat earlier points.'' \\

\texttt{14B-SEFL} &
``Clear and student-friendly while retaining depth.'' &
``Uses short sentences and connects advice to the rubric.'' &
``Groups comments by theme, which avoids repetition.'' \\
\bottomrule
\end{tabularx}
\caption{\textbf{Several Examples of Human Feedback.} We select several human annotator remarks that illustrate how SEFL tuning improves feedback quality compared with the original models.}
\label{tab:SEFL_vs_base_samples}
\end{table*}

\begin{figure*}[t]
    \centering
    \includegraphics[width=.9\linewidth]{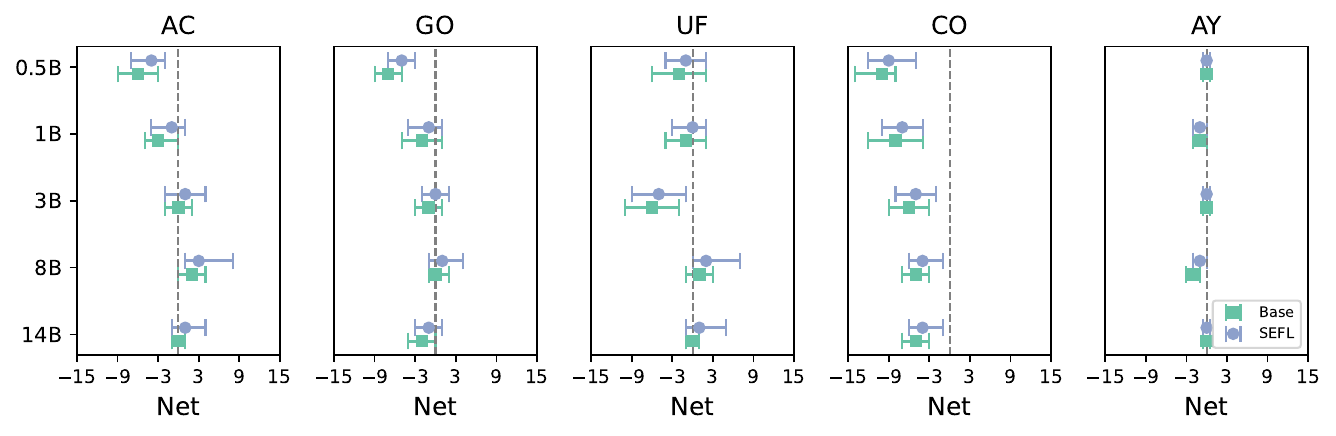}
    \caption{\textbf{Optional Rater Comments by Category.} AC = Actionability, GO = Goal-orientation, UF = User-friendliness, CO = Consistency, AY = Autonomy. Annotators were \emph{not} required to leave a comment; they did so mainly when a response stood out (usually for a problem). We also show the 95\% Wilson interval for the net balance; if it is not visible, it denotes zero comments. We show that SEFL-tuned models are getting more frequent positive (absolute) comments.}
    \label{fig:net_balances}
\end{figure*}

\begin{table*}[t]
\centering
\scriptsize

\rowcolors{1}{gray!20}{white}

\begin{tabularx}{\linewidth}{p{2.5cm} p{5.2cm} p{5.1cm} p{1.5cm}}
\toprule
\textbf{Aspect} & \textbf{Base Model} & \textbf{SEFL-tuned Model} & \textbf{Preferred} \\
\midrule
\textit{Consistency} &
``Feedback from model A fails to address key aspects of the answer, such as suddenly changing the name of the main character.'' &
``Not an answer but A properly identified it!'' &
SEFL \\
\midrule
\textit{Goal orientation} &
``In many cases, answers are shorter than required. Not reflected in feedback.'' &
``Finally, B finds that the answer is incomplete.'' &
SEFL \\
\midrule
\textit{Tone} &
``Model B: Harsh tone.'' &
``Both models are good, but model A is nicer in tone and actionability.'' &
SEFL \\
\midrule
\textit{User friendliness} &
``Model B is best, but is way too elaborate.'' &
``B is more clear and concise.'' &
SEFL \\
\midrule
\textit{Actionability} &
``Model A also provides partial solution.'' &
``Model B is not accurate and provides an answer instead of feedback.'' &
Base \\
\bottomrule
\end{tabularx}
\caption{\textbf{Representative Rater Comments.} We illustrate both strengths and weaknesses of SEFL-tuned models versus their base counterparts, based on the fine-grained criteria depicted in~\cref{subsec:analysis2}. The final column shows which draft the rater chose.}
\label{tab:qual_balance}
\end{table*}

\section{Discussion}
\subsection{Human Qualitative Insights}\label{subsec:analysis2}
In addition to the win rates in~\cref{fig:results}, our human annotators provided rich qualitative feedback on the model outputs, which we show in~\cref{tab:SEFL_vs_base_samples}. Generally, on the critical side, they noted that if a student's answer is too short or incomplete, neither model explicitly flags the missing details. More specifically, \qwenmini{} was praised for clarity and concision, whereas \llamasmall{} tended to repeat assignment details without offering actionable guidance. Annotators observed that \llamamini{} often gave more specific and constructive feedback but occasionally sounded harsh, while \llamamed{} sometimes overlooked key aspects. Overall, although \qwenmed{} achieved high win rates (94, 77, 81 across three annotators), these insights suggest that even top-performing models could improve in error detection, tone refinement, and contextual sensitivity.

To further quantify, in \cref{fig:net_balances}, we plot for each model the net balance of optional rater comments in five qualitative categories: Actionability (AC), Goal Orientation (GO), User Friendliness (UF), Consistency (CO), and Student Autonomy (AY)~\cite{carless2011developing,wiggins2012seven}. Squares denote the base models and circles the SEFL-tuned variants. Horizontal whiskers provide 95\% confidence intervals for the net balance using the bootstrap method. We compute these intervals on the proportion of positive remarks and then transform them to the net scale via 
$b = n \,\bigl(2p - 1\bigr)$,
where \(n\) is the total number of comments and \(p\) the positive proportion. Annotators added comments only when a response stood out, so the plot reveals both the direction and the strength of impressions. We show several examples of annotations in~\cref{tab:qual_balance}. The annotators also had the option to reject nonsensical feedback, which occurred around 12\% of the time, primarily with the smaller-sized models.

The two smallest models, \qwenmini{} and \llamamini{}, receive more negative than positive remarks on Consistency and Goal Orientation, matching earlier findings that they sometimes drift from the student answer or overlook core requirements.  \llamamed{} shows the only clearly positive balance in Actionability and User Friendliness, yet its interval for Consistency still lies below zero. \qwenmed{} gains more favorable notes on tone and clarity than the smaller models, but still shows a negative alignment gap. Regarding Student Autonomy, there are mostly neutral comments.
We show an example of the full output feedback of \qwenmini{} in \cref{fig:feedback-example} (\cref{app:example}).

\begin{table}[t]
    \centering
    \small
    \begin{tabular}{l r r r}
    \toprule
    \textbf{Method/Dataset}  & \textbf{J1}  & \textbf{J2} & \textbf{J3}  \\
    \midrule
    Book2Dial     & 43           & 37           & 27    \\
    SEFL          & \textbf{57}  & \textbf{63}  & \textbf{73}    \\
    \bottomrule
    \end{tabular}
    \caption{\textbf{SEFL versus Book2Dial.} We show the win rate between SEFL and Book2Dial with \qwenmed{} as the backbone model, evaluated by 3 LLM judges with criteria indicated in~\cref{subsec:eval}. In bold, we indicate the winning system. The 3 LLM judges are \texttt{gpt-4o} (J1), \texttt{claude-3.5-sonnet} (J2), and \texttt{deepseek-v3} (J3).}
    \label{tab:book2dial}
\end{table}

\subsection{LLM-as-a-Judge} 
We used LLM judges to rate the feedback generated by \sefl-tuned models against their untuned counterparts. 
This provides a scalable method for measuring feedback quality, thereby reducing the need for extensive human annotation. We let the LLM judges rate the same examples as the humans annotated.
As shown in~\cref{fig:results}, 3 out of 3 LLM judges consistently favored \sefl-tuned \qwenmini{}, \llamamini{}, and \llamasmall{}. 
We see it as a practical first step for large-scale feedback comparisons in educational contexts. We recommend supplementing LLM-based assessments with targeted human evaluations for more granular insights, possibly aligning more with instructional objectives or even fine-tuning LLMs with rubrics for better judgment of long-form text~\cite{kim2024prometheus}.

\subsection{Comparison to Prior Work}
The work that is most closely related to ours is Book2Dial~\cite{wang-etal-2024-book2dial}. It is a framework that turns textbooks into synthetic conversations between a student model and a teacher model. The student only sees high-level cues, such as section titles or key concepts, while the teacher has full access to the source passage, prompting a question-and-answer exchange that remains aligned with the book content. Instead, we focus on any assignment, not limited to textbooks.

To compare the two methods, we further fine-tune \qwenmed{} with the Book2Dial data in the same way as our SEFL-tuned version. We utilize the existing Book2Dial data, which consists of 889 dialogues, and preprocess the data to ensure that each conversation is one-turn, resulting in 5,300 conversation pairs. We then run the fine-tuned model over the same samples evaluated in Subsection~\ref{subsec:eval} and assess them with the same 3 judges and evaluation criteria.\footnote{At the time of writing, we ran out of funding for the human annotators and thus compare here only with LLM-as-a-Judge.} \cref{tab:book2dial} shows that 3 out of 3 judges prefer SEFL, yielding an average win rate of 64\%. The results confirm that SEFL produces higher quality feedback than the textbook-based dialogues in Book2Dial.

\section{Conclusion}
We introduced \sefl{}, a framework that simulates teacher$\rightarrow$student interactions via two-agent LLMs to generate synthetic data for fine-tuning smaller models. 
In this work, we aimed to explore how synthetic teacher-student interactions generated by LLMs can be leveraged for scalable feedback. We address this by demonstrating that these interactions, when generated through SEFL, can be used as high-quality fine-tuning data to enhance the feedback capabilities of smaller, more efficient language models.
This approach yields concise, context-sensitive feedback that often surpasses original instruction-tuned models under both LLM-as-a-judge and human evaluations. Yet, human insights remain indispensable for capturing nuances such as clarity and tone. SEFL provides a promising avenue for immediate, personalized feedback at scale, extending beyond the educational domain.

\section*{Limitations}
We acknowledge that \sefl{} relies on short-form, synthetically generated assignments and errors, which are not real student submissions, and this could have implications. Although this approach helps create large datasets, it risks producing feedback unaligned with authentic classroom contexts. Our evaluation also utilizes LLM-based judges, which introduces potential biases related to each judge's training data and objectives. Lastly, while we focused on short-answer tasks, longer or more domain-specific assignments may require specialized or more diverse synthetic data, which we leave to future work.

\section*{Ethical Statement}

The use of synthetic data provides an opportunity to train automated feedback systems without the constraints of privacy and consent that come from repurposing actual student assignments and teacher feedback as training data. However, it also raises questions about transparency and potential misuse~\cite{lindsay2024responsibledevelopmentautomatedstudent}. For instance, malicious actors could manipulate synthetic data to disseminate misleading or biased feedback, undermining trust in educational tools. Users may also mistake synthetic feedback for real, expert guidance. Moreover, automated feedback systems risk reinforcing biases if the underlying models are trained on skewed data. We believe educators and institutions should remain aware of these risks and incorporate human oversight to ensure that such systems \textit{complement}, rather than replace, genuine pedagogical engagement with real teachers.

\section*{Acknowledgements}
MZ, APD, JB, and EDL were supported by the research grant (VIL57392) from VILLUM FONDEN. We would like to thank the AAU-NLP group for helpful discussions and feedback on an earlier version of this article. 
MZ also received funding from the Danish Government to Danish Foundation Models (4378-00001B).
We acknowledge the Danish e-Infrastructure Cooperation for awarding this project access (No. 465001263; DeiC-AAU-N5-2024078 - H2-2024-18) to the LUMI supercomputer, owned by the EuroHPC Joint Undertaking, hosted by CSC (Finland) and the LUMI consortium through DeiC, Denmark.

\section*{Bibliographical References}\label{sec:reference}

\bibliographystyle{lrec2026-natbib}
\bibliography{lrec2026-example,anthology}

\begin{thebibliography}{62}
\expandafter\ifx\csname natexlab\endcsname\relax\def\natexlab#1{#1}\fi

\bibitem[{Alrajhi et~al.(2021)Alrajhi, Alamri, Pereira, and Cristea}]{alrajhi2021urgency}
Laila Alrajhi, Ahmed Alamri, Filipe~Dwan Pereira, and Alexandra~I Cristea. 2021.
\newblock Urgency analysis of learners’ comments: An automated intervention priority model for mooc.
\newblock In \emph{Intelligent Tutoring Systems: 17th International Conference, ITS 2021, Virtual Event, June 7--11, 2021, Proceedings 17}, pages 148--160. Springer.

\bibitem[{Aslan et~al.(2019)Aslan, Alyuz, Tanriover, Mete, Okur, D'Mello, and Arslan~Esme}]{aslan2019investigating}
Sinem Aslan, Nese Alyuz, Cagri Tanriover, Sinem~E Mete, Eda Okur, Sidney~K D'Mello, and Asli Arslan~Esme. 2019.
\newblock Investigating the impact of a real-time, multimodal student engagement analytics technology in authentic classrooms.
\newblock In \emph{Proceedings of the 2019 chi conference on human factors in computing systems}, pages 1--12.

\bibitem[{AtlaAI(2025)}]{judge-arena}
AtlaAI. 2025.
\newblock Judge arena.
\newblock \url{https://huggingface.co/spaces/AtlaAI/judge-arena}.
\newblock [Online; accessed 8-April-2025].

\bibitem[{Bauer et~al.(2023)Bauer, Greisel, Kuznetsov, Berndt, Kollar, Dresel, Fischer, and Fischer}]{bauer2023using}
Elisabeth Bauer, Martin Greisel, Ilia Kuznetsov, Markus Berndt, Ingo Kollar, Markus Dresel, Martin~R Fischer, and Frank Fischer. 2023.
\newblock Using natural language processing to support peer-feedback in the age of artificial intelligence: A cross-disciplinary framework and a research agenda.
\newblock \emph{British Journal of Educational Technology}, 54(5):1222--1245.

\bibitem[{Botelho et~al.(2023)Botelho, Baral, Erickson, Benachamardi, and Heffernan}]{botelho2023leveraging}
Anthony Botelho, Sami Baral, John~A Erickson, Priyanka Benachamardi, and Neil~T Heffernan. 2023.
\newblock Leveraging natural language processing to support automated assessment and feedback for student open responses in mathematics.
\newblock \emph{Journal of computer assisted learning}, 39(3):823--840.

\bibitem[{Carless et~al.(2011)Carless, Salter, Yang, and Lam}]{carless2011developing}
David Carless, Diane Salter, Min Yang, and Joy Lam. 2011.
\newblock Developing sustainable feedback practices.
\newblock \emph{Studies in higher education}, 36(4):395--407.

\bibitem[{Chen et~al.(2023)Chen, Wang, Jiang, Shi, and Xu}]{chen-etal-2023-exploring-use}
Yi~Chen, Rui Wang, Haiyun Jiang, Shuming Shi, and Ruifeng Xu. 2023.
\newblock \href {https://doi.org/10.18653/v1/2023.findings-ijcnlp.32} {Exploring the use of large language models for reference-free text quality evaluation: An empirical study}.
\newblock In \emph{Findings of the Association for Computational Linguistics: IJCNLP-AACL 2023 (Findings)}, pages 361--374, Nusa Dua, Bali. Association for Computational Linguistics.

\bibitem[{Cohen(1960)}]{cohen1960coefficient}
Jacob Cohen. 1960.
\newblock A coefficient of agreement for nominal scales.
\newblock \emph{Educational and psychological measurement}, 20(1):37--46.

\bibitem[{Conole and Oliver(2006)}]{conole2006contemporary}
Grainne Conole and Martin Oliver. 2006.
\newblock \emph{Contemporary perspectives in e-learning research}.
\newblock Routledge London.

\bibitem[{Costello and Crane(2013)}]{costello2013technologies}
Jane Costello and Daph Crane. 2013.
\newblock Technologies for learner-centered feedback.
\newblock \emph{Open Praxis}, 5(3):217--225.

\bibitem[{Demszky and Hill(2023)}]{demszky-hill-2023-ncte}
Dorottya Demszky and Heather Hill. 2023.
\newblock \href {https://doi.org/10.18653/v1/2023.bea-1.44} {The {NCTE} transcripts: A dataset of elementary math classroom transcripts}.
\newblock In \emph{Proceedings of the 18th Workshop on Innovative Use of NLP for Building Educational Applications (BEA 2023)}, pages 528--538, Toronto, Canada. Association for Computational Linguistics.

\bibitem[{Falk et~al.(2025)Falk, Chen, Rafner, Zhang, Bjerva, and Nolte}]{falkandchen2025how}
Jeanette Falk, Yiyi Chen, Janet Rafner, Mike Zhang, Johannes Bjerva, and Alexander Nolte. 2025.
\newblock \href {https://doi.org/10.1145/3706598.3713447} {How do hackathons foster creativity? towards ai collaborative evaluation of creativity at scale}.
\newblock In \emph{Proceedings of the 2025 CHI Conference on Human Factors in Computing Systems}, CHI '25. Association for Computing Machinery.

\bibitem[{Ferguson(2011)}]{ferguson2011student}
Peter Ferguson. 2011.
\newblock Student perceptions of quality feedback in teacher education.
\newblock \emph{Assessment \& evaluation in higher education}, 36(1):51--62.

\bibitem[{Fischer et~al.(2020)Fischer, Pardos, Baker, Williams, Smyth, Yu, Slater, Baker, and Warschauer}]{fischer2020mining}
Christian Fischer, Zachary~A Pardos, Ryan~Shaun Baker, Joseph~Jay Williams, Padhraic Smyth, Renzhe Yu, Stefan Slater, Rachel Baker, and Mark Warschauer. 2020.
\newblock Mining big data in education: Affordances and challenges.
\newblock \emph{Review of Research in Education}, 44(1):130--160.

\bibitem[{Gilardi et~al.(2023)Gilardi, Alizadeh, and Kubli}]{gilardi2023chatgpt}
Fabrizio Gilardi, Meysam Alizadeh, and Ma{\"e}l Kubli. 2023.
\newblock Chatgpt outperforms crowd workers for text-annotation tasks.
\newblock \emph{Proceedings of the National Academy of Sciences}, 120(30):e2305016120.

\bibitem[{Gu et~al.(2024)Gu, Jiang, Shi, Tan, Zhai, Xu, Li, Shen, Ma, Liu, Wang, and Guo}]{gu2024surveyllmasajudge}
Jiawei Gu, Xuhui Jiang, Zhichao Shi, Hexiang Tan, Xuehao Zhai, Chengjin Xu, Wei Li, Yinghan Shen, Shengjie Ma, Honghao Liu, Yuanzhuo Wang, and Jian Guo. 2024.
\newblock \href {http://arxiv.org/abs/2411.15594} {A survey on llm-as-a-judge}.

\bibitem[{Guerraoui et~al.(2023)Guerraoui, Reisert, Inoue, Mim, Singh, Choi, Robbani, Naito, Wang, and Inui}]{guerraoui-etal-2023-teach}
Camelia Guerraoui, Paul Reisert, Naoya Inoue, Farjana~Sultana Mim, Keshav Singh, Jungmin Choi, Irfan Robbani, Shoichi Naito, Wenzhi Wang, and Kentaro Inui. 2023.
\newblock \href {https://doi.org/10.18653/v1/2023.argmining-1.3} {Teach me how to argue: A survey on {NLP} feedback systems in argumentation}.
\newblock In \emph{Proceedings of the 10th Workshop on Argument Mining}, pages 19--34, Singapore. Association for Computational Linguistics.

\bibitem[{Hattie(2008)}]{hattie2008visible}
John Hattie. 2008.
\newblock \emph{Visible learning: A synthesis of over 800 meta-analyses relating to achievement}.
\newblock routledge.

\bibitem[{Huang et~al.(2024)Huang, Kwak, Park, and An}]{huang-etal-2024-chatgpt}
Fan Huang, Haewoon Kwak, Kunwoo Park, and Jisun An. 2024.
\newblock \href {https://aclanthology.org/2024.lrec-main.277/} {{C}hat{GPT} rates natural language explanation quality like humans: But on which scales?}
\newblock In \emph{Proceedings of the 2024 Joint International Conference on Computational Linguistics, Language Resources and Evaluation (LREC-COLING 2024)}, pages 3111--3132, Torino, Italia. ELRA and ICCL.

\bibitem[{Hurst et~al.(2024)Hurst, Lerer, Goucher, Perelman, Ramesh, Clark, Ostrow, Welihinda, Hayes, Radford et~al.}]{hurst2024gpt}
Aaron Hurst, Adam Lerer, Adam~P Goucher, Adam Perelman, Aditya Ramesh, Aidan Clark, AJ~Ostrow, Akila Welihinda, Alan Hayes, Alec Radford, et~al. 2024.
\newblock Gpt-4o system card.
\newblock \emph{arXiv preprint arXiv:2410.21276}.

\bibitem[{Ke and Ng(2019)}]{ke2019automated}
Zixuan Ke and Vincent Ng. 2019.
\newblock Automated essay scoring: A survey of the state of the art.
\newblock In \emph{IJCAI}, volume~19, pages 6300--6308.

\bibitem[{Kim et~al.(2022)Kim, Kim, Yoo, and Kang}]{kim-etal-2022-generating}
Gangwoo Kim, Sungdong Kim, Kang~Min Yoo, and Jaewoo Kang. 2022.
\newblock \href {https://doi.org/10.18653/v1/2022.emnlp-main.151} {Generating information-seeking conversations from unlabeled documents}.
\newblock In \emph{Proceedings of the 2022 Conference on Empirical Methods in Natural Language Processing}, pages 2362--2378, Abu Dhabi, United Arab Emirates. Association for Computational Linguistics.

\bibitem[{Kim et~al.(2024)Kim, Suk, Longpre, Lin, Shin, Welleck, Neubig, Lee, Lee, and Seo}]{kim2024prometheus}
Seungone Kim, Juyoung Suk, Shayne Longpre, Bill~Yuchen Lin, Jamin Shin, Sean Welleck, Graham Neubig, Moontae Lee, Kyungjae Lee, and Minjoon Seo. 2024.
\newblock Prometheus 2: An open source language model specialized in evaluating other language models.
\newblock \emph{arXiv preprint arXiv:2405.01535}.

\bibitem[{Kocmi and Federmann(2023)}]{kocmi-federmann-2023-large}
Tom Kocmi and Christian Federmann. 2023.
\newblock \href {https://aclanthology.org/2023.eamt-1.19/} {Large language models are state-of-the-art evaluators of translation quality}.
\newblock In \emph{Proceedings of the 24th Annual Conference of the European Association for Machine Translation}, pages 193--203, Tampere, Finland. European Association for Machine Translation.

\bibitem[{Kwon et~al.(2024)Kwon, Kim, Park, Lee, and Kim}]{kwon2024biped}
Soonwoo Kwon, Sojung Kim, Minju Park, Seunghyun Lee, and Kyuseok Kim. 2024.
\newblock Biped: Pedagogically informed tutoring system for esl education.
\newblock \emph{arXiv preprint arXiv:2406.03486}.

\bibitem[{Lambert et~al.(2024)Lambert, Pyatkin, Morrison, Miranda, Lin, Chandu, Dziri, Kumar, Zick, Choi et~al.}]{lambert2024rewardbench}
Nathan Lambert, Valentina Pyatkin, Jacob Morrison, LJ~Miranda, Bill~Yuchen Lin, Khyathi Chandu, Nouha Dziri, Sachin Kumar, Tom Zick, Yejin Choi, et~al. 2024.
\newblock Rewardbench: Evaluating reward models for language modeling.
\newblock \emph{arXiv preprint arXiv:2403.13787}.

\bibitem[{Landis and Koch(1977)}]{landis1977measurement}
J~Richard Landis and Gary~G Koch. 1977.
\newblock The measurement of observer agreement for categorical data.
\newblock \emph{biometrics}, pages 159--174.

\bibitem[{Li et~al.(2023)Li, Hammoud, Itani, Khizbullin, and Ghanem}]{li2023camel}
Guohao Li, Hasan Hammoud, Hani Itani, Dmitrii Khizbullin, and Bernard Ghanem. 2023.
\newblock Camel: Communicative agents for" mind" exploration of large language model society.
\newblock \emph{Advances in Neural Information Processing Systems}, 36:51991--52008.

\bibitem[{Liang et~al.(2024)Liang, Zhang, Cao, Wang, Ding, Yang, Vodrahalli, He, Smith, Yin et~al.}]{liang2024can}
Weixin Liang, Yuhui Zhang, Hancheng Cao, Binglu Wang, Daisy~Yi Ding, Xinyu Yang, Kailas Vodrahalli, Siyu He, Daniel~Scott Smith, Yian Yin, et~al. 2024.
\newblock Can large language models provide useful feedback on research papers? a large-scale empirical analysis.
\newblock \emph{NEJM AI}, 1(8):AIoa2400196.

\bibitem[{Lindsay et~al.(2024)Lindsay, Zhang, Johri, and Bjerva}]{lindsay2024responsibledevelopmentautomatedstudent}
Euan~D Lindsay, Mike Zhang, Aditya Johri, and Johannes Bjerva. 2024.
\newblock \href {http://arxiv.org/abs/2308.15334} {The responsible development of automated student feedback with generative ai}.

\bibitem[{Liu et~al.(2024{\natexlab{a}})Liu, Feng, Xue, Wang, Wu, Lu, Zhao, Deng, Zhang, Ruan et~al.}]{liu2024deepseek}
Aixin Liu, Bei Feng, Bing Xue, Bingxuan Wang, Bochao Wu, Chengda Lu, Chenggang Zhao, Chengqi Deng, Chenyu Zhang, Chong Ruan, et~al. 2024{\natexlab{a}}.
\newblock Deepseek-v3 technical report.
\newblock \emph{arXiv preprint arXiv:2412.19437}.

\bibitem[{Liu et~al.(2024{\natexlab{b}})Liu, Huang, Xiao, Sha, Wu, Liu, Wang, and Chen}]{liu2024socraticlm}
Jiayu Liu, Zhenya Huang, Tong Xiao, Jing Sha, Jinze Wu, Qi~Liu, Shijin Wang, and Enhong Chen. 2024{\natexlab{b}}.
\newblock Socraticlm: exploring socratic personalized teaching with large language models.
\newblock \emph{Advances in Neural Information Processing Systems}, 37:85693--85721.

\bibitem[{Liu et~al.(2023)Liu, Iter, Xu, Wang, Xu, and Zhu}]{liu-etal-2023-g}
Yang Liu, Dan Iter, Yichong Xu, Shuohang Wang, Ruochen Xu, and Chenguang Zhu. 2023.
\newblock \href {https://doi.org/10.18653/v1/2023.emnlp-main.153} {{G}-eval: {NLG} evaluation using gpt-4 with better human alignment}.
\newblock In \emph{Proceedings of the 2023 Conference on Empirical Methods in Natural Language Processing}, pages 2511--2522, Singapore. Association for Computational Linguistics.

\bibitem[{Liu et~al.(2024{\natexlab{c}})Liu, Yin, Lin, and Chen}]{liu2024personality}
Zhengyuan Liu, Stella~Xin Yin, Geyu Lin, and Nancy~F Chen. 2024{\natexlab{c}}.
\newblock Personality-aware student simulation for conversational intelligent tutoring systems.
\newblock \emph{arXiv preprint arXiv:2404.06762}.

\bibitem[{Loshchilov and Hutter(2019)}]{loshchilov2018decoupled}
Ilya Loshchilov and Frank Hutter. 2019.
\newblock \href {https://openreview.net/forum?id=Bkg6RiCqY7} {Decoupled weight decay regularization}.
\newblock In \emph{7th International Conference on Learning Representations, {ICLR} 2019, New Orleans, LA, USA, May 6-9, 2019}. OpenReview.net.

\bibitem[{Lozhkov et~al.(2024)Lozhkov, Ben~Allal, von Werra, and Wolf}]{lozhkov2024fineweb-edu}
Anton Lozhkov, Loubna Ben~Allal, Leandro von Werra, and Thomas Wolf. 2024.
\newblock \href {https://doi.org/10.57967/hf/2497} {Fineweb-edu}.

\bibitem[{Nair et~al.(2024)Nair, Tan, Su, Gere, Wang, and Wang}]{nair-etal-2024-closing}
Inderjeet~Jayakumar Nair, Jiaye Tan, Xiaotian Su, Anne Gere, Xu~Wang, and Lu~Wang. 2024.
\newblock \href {https://doi.org/10.18653/v1/2024.emnlp-main.928} {Closing the loop: Learning to generate writing feedback via language model simulated student revisions}.
\newblock In \emph{Proceedings of the 2024 Conference on Empirical Methods in Natural Language Processing}, pages 16636--16657, Miami, Florida, USA. Association for Computational Linguistics.

\bibitem[{Naismith et~al.(2023)Naismith, Mulcaire, and Burstein}]{naismith-etal-2023-automated}
Ben Naismith, Phoebe Mulcaire, and Jill Burstein. 2023.
\newblock \href {https://doi.org/10.18653/v1/2023.bea-1.32} {Automated evaluation of written discourse coherence using {GPT}-4}.
\newblock In \emph{Proceedings of the 18th Workshop on Innovative Use of NLP for Building Educational Applications (BEA 2023)}, pages 394--403, Toronto, Canada. Association for Computational Linguistics.

\bibitem[{Nicol(2007)}]{nicol2007assessment}
David Nicol. 2007.
\newblock E-assessment by design: using multiple-choice tests to good effect.
\newblock \emph{Journal of Further and higher Education}, 31(1):53--64.

\bibitem[{Rafailov et~al.(2023)Rafailov, Sharma, Mitchell, Manning, Ermon, and Finn}]{rafailov2023direct}
Rafael Rafailov, Archit Sharma, Eric Mitchell, Christopher~D Manning, Stefano Ermon, and Chelsea Finn. 2023.
\newblock Direct preference optimization: Your language model is secretly a reward model.
\newblock \emph{Advances in neural information processing systems}, 36:53728--53741.

\bibitem[{Ramesh and Sanampudi(2022)}]{ramesh2022automated}
Dadi Ramesh and Suresh~Kumar Sanampudi. 2022.
\newblock An automated essay scoring systems: a systematic literature review.
\newblock \emph{Artificial Intelligence Review}, 55(3):2495--2527.

\bibitem[{Rooein and Hovy(2024)}]{rooein2024conversations}
Donya Rooein and Dirk Hovy. 2024.
\newblock Conversations as a source for teaching scientific concepts at different education levels.
\newblock \emph{arXiv preprint arXiv:2404.10475}.

\bibitem[{Ross and Andreas(2024)}]{ross-andreas-2024-toward}
Alexis Ross and Jacob Andreas. 2024.
\newblock \href {https://doi.org/10.18653/v1/2024.acl-long.718} {Toward in-context teaching: Adapting examples to students' misconceptions}.
\newblock In \emph{Proceedings of the 62nd Annual Meeting of the Association for Computational Linguistics (Volume 1: Long Papers)}, pages 13283--13310, Bangkok, Thailand. Association for Computational Linguistics.

\bibitem[{Saito et~al.(2023)Saito, Wachi, Wataoka, and Akimoto}]{saito2023verbosity}
Keita Saito, Akifumi Wachi, Koki Wataoka, and Youhei Akimoto. 2023.
\newblock Verbosity bias in preference labeling by large language models.
\newblock \emph{arXiv preprint arXiv:2310.10076}.

\bibitem[{Schwarz et~al.(2018)Schwarz, Prusak, Swidan, Livny, Gal, and Segal}]{schwarz2018orchestrating}
Baruch~B Schwarz, Naomi Prusak, Osama Swidan, Adva Livny, Kobi Gal, and Avi Segal. 2018.
\newblock Orchestrating the emergence of conceptual learning: A case study in a geometry class.
\newblock \emph{International Journal of Computer-Supported Collaborative Learning}, 13:189--211.

\bibitem[{Sonkar et~al.(2024)Sonkar, Ni, Chaudhary, and Baraniuk}]{sonkar2024pedagogical}
Shashank Sonkar, Kangqi Ni, Sapana Chaudhary, and Richard~G Baraniuk. 2024.
\newblock Pedagogical alignment of large language models.
\newblock \emph{arXiv preprint arXiv:2402.05000}.

\bibitem[{Stahl et~al.(2024)Stahl, Biermann, Nehring, and Wachsmuth}]{stahl-etal-2024-exploring}
Maja Stahl, Leon Biermann, Andreas Nehring, and Henning Wachsmuth. 2024.
\newblock \href {https://aclanthology.org/2024.bea-1.23/} {Exploring {LLM} prompting strategies for joint essay scoring and feedback generation}.
\newblock In \emph{Proceedings of the 19th Workshop on Innovative Use of NLP for Building Educational Applications (BEA 2024)}, pages 283--298, Mexico City, Mexico. Association for Computational Linguistics.

\bibitem[{Suresh et~al.(2022)Suresh, Jacobs, Harty, Perkoff, Martin, and Sumner}]{suresh-etal-2022-talkmoves}
Abhijit Suresh, Jennifer Jacobs, Charis Harty, Margaret Perkoff, James~H. Martin, and Tamara Sumner. 2022.
\newblock \href {https://aclanthology.org/2022.lrec-1.497/} {The {T}alk{M}oves dataset: K-12 mathematics lesson transcripts annotated for teacher and student discursive moves}.
\newblock In \emph{Proceedings of the Thirteenth Language Resources and Evaluation Conference}, pages 4654--4662, Marseille, France. European Language Resources Association.

\bibitem[{Tan et~al.(2024)Tan, Zhuang, Montgomery, Tang, Cuadron, Wang, Popa, and Stoica}]{tan2024judgebench}
Sijun Tan, Siyuan Zhuang, Kyle Montgomery, William~Y Tang, Alejandro Cuadron, Chenguang Wang, Raluca~Ada Popa, and Ion Stoica. 2024.
\newblock Judgebench: A benchmark for evaluating llm-based judges.
\newblock \emph{arXiv preprint arXiv:2410.12784}.

\bibitem[{T{\"o}rnberg(2023)}]{tornberg2023chatgpt}
Petter T{\"o}rnberg. 2023.
\newblock Chatgpt-4 outperforms experts and crowd workers in annotating political twitter messages with zero-shot learning.
\newblock \emph{arXiv preprint arXiv:2304.06588}.

\bibitem[{Verga et~al.(2024)Verga, Hofstatter, Althammer, Su, Piktus, Arkhangorodsky, Xu, White, and Lewis}]{verga2024replacing}
Pat Verga, Sebastian Hofstatter, Sophia Althammer, Yixuan Su, Aleksandra Piktus, Arkady Arkhangorodsky, Minjie Xu, Naomi White, and Patrick Lewis. 2024.
\newblock Replacing judges with juries: Evaluating llm generations with a panel of diverse models.
\newblock \emph{arXiv preprint arXiv:2404.18796}.

\bibitem[{Wang et~al.(2024{\natexlab{a}})Wang, Macina, Daheim, Pal~Chowdhury, and Sachan}]{wang-etal-2024-book2dial}
Junling Wang, Jakub Macina, Nico Daheim, Sankalan Pal~Chowdhury, and Mrinmaya Sachan. 2024{\natexlab{a}}.
\newblock \href {https://doi.org/10.18653/v1/2024.findings-acl.578} {{B}ook2{D}ial: Generating teacher student interactions from textbooks for cost-effective development of educational chatbots}.
\newblock In \emph{Findings of the Association for Computational Linguistics: ACL 2024}, pages 9707--9731, Bangkok, Thailand. Association for Computational Linguistics.

\bibitem[{Wang and Demszky(2024)}]{wang-demszky-2024-edu}
Rose Wang and Dorottya Demszky. 2024.
\newblock \href {https://doi.org/10.18653/v1/2024.naacl-demo.6} {Edu-{C}onvo{K}it: An open-source library for education conversation data}.
\newblock In \emph{Proceedings of the 2024 Conference of the North American Chapter of the Association for Computational Linguistics: Human Language Technologies (Volume 3: System Demonstrations)}, pages 61--69, Mexico City, Mexico. Association for Computational Linguistics.

\bibitem[{Wang et~al.(2024{\natexlab{b}})Wang, Ribeiro, Robinson, Loeb, and Demszky}]{wang2024tutor}
Rose~E Wang, Ana~T Ribeiro, Carly~D Robinson, Susanna Loeb, and Dora Demszky. 2024{\natexlab{b}}.
\newblock Tutor copilot: A human-ai approach for scaling real-time expertise.
\newblock \emph{arXiv preprint arXiv:2410.03017}.

\bibitem[{Wang et~al.(2024{\natexlab{c}})Wang, Xu, Li, Zhang, Liang, Tang, Yu, and Wen}]{wang2024large}
Shen Wang, Tianlong Xu, Hang Li, Chaoli Zhang, Joleen Liang, Jiliang Tang, Philip~S Yu, and Qingsong Wen. 2024{\natexlab{c}}.
\newblock Large language models for education: A survey and outlook.
\newblock \emph{arXiv preprint arXiv:2403.18105}.

\bibitem[{Wiggins(2012)}]{wiggins2012seven}
Grant Wiggins. 2012.
\newblock Seven keys to effective feedback.
\newblock \emph{Feedback}, 70(1):10--16.

\bibitem[{Wu et~al.(2023)Wu, Bansal, Zhang, Wu, Zhang, Zhu, Li, Jiang, Zhang, and Wang}]{wu2023autogen}
Qingyun Wu, Gagan Bansal, Jieyu Zhang, Yiran Wu, Shaokun Zhang, Erkang Zhu, Beibin Li, Li~Jiang, Xiaoyun Zhang, and Chi Wang. 2023.
\newblock Autogen: Enabling next-gen llm applications via multi-agent conversation framework.
\newblock \emph{arXiv preprint arXiv:2308.08155}.

\bibitem[{Yun et~al.(2024)Yun, Hicke, Olson, and Demszky}]{yun2024enhancing}
Joy Yun, Yann Hicke, Mariah Olson, and Dorottya Demszky. 2024.
\newblock Enhancing tutoring effectiveness through automated feedback: Preliminary findings from a pilot randomized controlled trial on sat tutoring.
\newblock In \emph{Proceedings of the Eleventh ACM Conference on Learning@ Scale}, pages 422--426.

\bibitem[{Zhang et~al.(2025)Zhang, Lindsay, Quitzau, and Bjerva}]{2babc566a8034738ac303e9ba612df14}
Mike Zhang, Euan Lindsay, Maj-Britt Quitzau, and Johannes Bjerva. 2025.
\newblock Scaling course evaluations with large language models: Semester-level digestible student feedback for program leaders.
\newblock In \emph{Proceedings of the 53rd Annual Conference of the European Society for Engineering Education (SEFI 2025)}.

\bibitem[{Zhang et~al.(2024)Zhang, Lindsay, Thorbensen, Poulsen, and Bjerva}]{19b02384b88f4404b0a2f5d5eec1207f}
Mike Zhang, Euan Lindsay, Frederik~Bode Thorbensen, Danny~B{\o}gsted Poulsen, and Johannes Bjerva. 2024.
\newblock \href {https://doi.org/10.48550/arXiv.2407.01274} {Leveraging large language models for actionable course evaluation student feedback to lecturers}.
\newblock In \emph{Proceedings of the 52nd Annual Conference of the European Society for Engineering Education (SEFI)}, pages 1089--1098.

\bibitem[{Zhang et~al.(2020)Zhang, Kishore, Wu, Weinberger, and Artzi}]{Zhang2020BERTScore}
Tianyi Zhang, Varsha Kishore, Felix Wu, Kilian~Q. Weinberger, and Yoav Artzi. 2020.
\newblock \href {https://openreview.net/forum?id=SkeHuCVFDr} {Bertscore: Evaluating text generation with {BERT}}.
\newblock In \emph{8th International Conference on Learning Representations, {ICLR} 2020, Addis Ababa, Ethiopia, April 26-30, 2020}. OpenReview.net.

\bibitem[{Zheng et~al.(2024)Zheng, Chiang, Sheng, Zhuang, Wu, Zhuang, Lin, Li, Li, Xing et~al.}]{zheng2024judging}
Lianmin Zheng, Wei-Lin Chiang, Ying Sheng, Siyuan Zhuang, Zhanghao Wu, Yonghao Zhuang, Zi~Lin, Zhuohan Li, Dacheng Li, Eric Xing, et~al. 2024.
\newblock Judging llm-as-a-judge with mt-bench and chatbot arena.
\newblock \emph{Advances in Neural Information Processing Systems}, 36.

\end{thebibliography}

\appendix
\section*{Appendix}

\section{Prompts}
\label{sec:prompts}
In~\cref{fig:prompt}, we show the prompts that we give to the agent models. Additionally, in~\cref{fig:prompt-llm}, we show the LLM-as-a-judge that we give to the judge models.

\begin{figure*}[t]
\centering
\begin{tcolorbox}[title=Prompts for Agent-based Educational Feedback Loop, promptstyle]
\lstset{
    basicstyle=\normalfont\sffamily\footnotesize,
    breaklines=true,
    frame=none,
    columns=fullflexible,
}
\begin{lstlisting}[linewidth=\linewidth]
##########################
### Student System Prompt ###
##########################

You are a diligent student who solves all assignments efficiently. Your key traits are: 
1. Direct and Concise Answers: Answer questions directly and concisely; use appropriate academic language. 
2. Show Your Work: Demonstrate your problem-solving process; provide step-by-step solutions when necessary. 
3. Encourage Learning: Focus on assisting with academic tasks; promote understanding through your answers. 
4. Intentional Mistakes: Make some obvious mistakes that the teacher can give feedback on; ensure mistakes are explicit and noticeable. 
5. Response Format: When responding to the teacher's assignment, give your answer and make explicit errors in your answer in valid JSON Lines (JSONL) format without any additional text, using the structure: {'answer': 'Your answer here', 'error_1': 'Description of the first mistake', 'error_2': 'Description of the second mistake'}. Do not write anything else.


##########################
### Teacher System Prompt ###
##########################

You are a skilled teacher specializing in creating concise, effective assignments and providing constructive, targeted feedback. Your key responsibilities are: 
1. Assignment Creation: Create short, clear assignments across various subjects; provide brief, focused instructions. 
2. Feedback Provision: Offer constructive feedback on completed work; explain concepts succinctly when needed; do not give grades, only feedback for each mistake. 
3. Encouragement and Adaptation: Encourage critical thinking and creativity; adapt to different learning styles and levels. 
4. Response Format: When creating an assignment, give your answer in valid JSON format using {'assignment': 'Your assignment text here', 'task': 'Specific task instructions here'}; when providing feedback on a student's reply, respond in valid JSONL format with {'answer': 'Your global feedback here', 'feedback_1': 'Feedback on the first mistake', 'feedback_2': 'Feedback on the second mistake'}. Do not write anything else. Your goal is to facilitate learning through well-designed tasks and helpful guidance.


######################
### Initial User Prompt ###
######################

{Fineweb-Edu Text Example}
\n\n
Create a short and concise one-question higher education level assignment given the text, be creative. Give your answer in valid jsonl format: {assignment: <text>, task_1: <text>, task_2: <text>, ...}. Do not write anything else.

\end{lstlisting}
\end{tcolorbox}
    \caption{\textbf{Prompt for Generating Synthetic Teacher$\rightarrow$Student Feedback Loops.} We show the prompt we use for the agentic setting.}
    \label{fig:prompt}
\end{figure*}

\begin{figure*}[t]
\centering
\begin{tcolorbox}[title=Prompt LLM-as-a-judge, promptstyle]
\lstset{
    basicstyle=\normalfont\sffamily\footnotesize,
    breaklines=true,
    frame=none,
    columns=fullflexible,
}
\begin{lstlisting}[linewidth=\linewidth]
##################
### Judge Prompt ###
##################

You are tasked with evaluating assignment feedback provided by two different models (Model A and Model B). As an objective evaluator, follow these steps: 
1. Analysis Criteria: 
- Accuracy: Does the feedback directly address specific strengths and weaknesses without unnecessary elaboration? 
- Actionability: Are suggestions clear, specific, and implementable without being overly prescriptive? 
- Conciseness: Is the feedback brief and focused while remaining meaningful? 
- Tone: Does the feedback maintain efficiency while being constructive? 
2. Evaluation Process: 
- First, review the original assignment task carefully 
- Then examine both Model A's and Model B's feedback responses 
- Compare them against the above criteria 
- Prioritize focused, efficient feedback over exhaustive detail 
3. Scoring Rules: 
- Responses should not include numerical grades 
- Feedback must be concise and directly related to the student's work 
- Each point should be essential and identify specific aspects of the response 
- Avoid unnecessary categorization and theoretical benefits 
4. Output Format: 
- Respond with a single character: 'A' or 'B' 
- Choose the model that provides more targeted, efficient feedback 
- Do not provide any additional explanation or commentary 
- Your response must contain exactly one character.

Assignment Prompt:
{prompt}

Model A feedback:
{model_a_feedback}

Model B feedback:
{model_b_feedback}

Which is better? Please respond with a single character: A or B."

\end{lstlisting}
\end{tcolorbox}
    \caption{\textbf{Prompt for LLM-as-a-Judge.} We show the prompt that we use for each LLM-as-a-Judge.}
    \label{fig:prompt-llm}
\end{figure*}

\section{Human Evaluation Guidelines}
\label{app:humaneval}
In~\cref{tab:annotation_guidelines}, we show the annotation guidelines for the human raters to rate the model feedback. The annotators were also instructed that the data will be made publicly available.

\begin{table*}[t]
\centering
\scriptsize
\renewcommand{\arraystretch}{1.5}
\begin{tabularx}{\textwidth}{|p{3cm}|X|}
\hline
\rowcolor[HTML]{EFEFEF} \textbf{Section} & \textbf{Details} \\ \hline

\textbf{Overview} & 
Your task is to evaluate pairs of feedback responses (Model A and Model B) given to student assignments. You will select which model provides better feedback according to specific criteria.
\newline
\textbf{Key Principles:}
\begin{itemize}
    \item Focus on efficiency and specificity.
    \item Value concise, meaningful feedback over lengthy explanations.
    \item Prioritize direct, actionable suggestions.
    \item Consider both content and delivery.
\end{itemize}
Remember to take breaks; I suggest spending a maximum of 10 minutes per row.
\\ \hline

\textbf{Sheet Information} & 
In the table, pick the one you got assigned. You will see 7 columns and need to fill in columns C and F:
\begin{itemize}
    \item \textbf{Appendix\_assignment:} What the large language model saw when generating an assignment with a possible answer.
    \item \textbf{Assignment:} What the model generated as an assignment and answered.
    \item \textbf{Model A:} Feedback generated by Model A.
    \item \textbf{Model B:} Feedback generated by Model B.
    \item \textbf{Which is better?} The most important part is to evaluate both feedback responses and determine which one is better, based on the assignment and answer.
    \item \textbf{Comments:} Leave comments if needed.
\end{itemize}
\\ \hline

\textbf{Evaluation Criteria} & 
\textbf{Accuracy:} Does the feedback address specific strengths and weaknesses? Are comments relevant to the student work? Is the critique substantive rather than superficial? 
\newline
\textbf{Actionability:} Are suggestions clear and specific? Can students easily understand what to improve? Are recommendations implementable? 
\newline
\textbf{Conciseness:} Is the feedback brief while remaining meaningful? Does it avoid unnecessary elaboration? Is there minimal redundancy? 
\newline
\textbf{Tone:} Is the feedback constructive while being efficient? Does it balance recognition with criticism? Is the language professional? 
\\ \hline

\textbf{Format} & 
\textbf{Preferred Feedback Style:}
\begin{itemize}
    \item Shows good understanding of the concept.
    \item Uses specific examples from the text to support arguments.
    \item Addresses the main question directly.
\end{itemize}
\textbf{Less Preferred Feedback Style:}
\begin{itemize}
    \item Generalized or vague feedback.
    \item Overly verbose or structured responses.
    \item Focuses on theoretical completeness rather than practical advice.
\end{itemize}
\\ \hline

\textbf{Scoring and Pitfalls} & 
\textbf{Scoring:}
\begin{enumerate}
    \item Read the original assignment carefully.
    \item Review both feedback responses.
    \item Evaluate against the criteria.
    \item Select the model that better aligns with the criteria as ``A'' or ``B.''
\end{enumerate}
\textbf{Pitfalls:}
\begin{itemize}
    \item Avoid preferring longer feedback just because it’s lengthy.
    \item Do not choose feedback that only lists general principles.
    \item Avoid letting formatting alone affect your choice.
\end{itemize}
\\ \hline
\end{tabularx}
\caption{Human Annotation Guidelines for Evaluating Assignment Feedback.}
\label{tab:annotation_guidelines}
\end{table*}

\section{Feedback Example}\label{app:example}
In~\cref{fig:feedback-example}, we show an example of the feedback of \qwenmini{} tuned on \sefl{} and out-of-the-box.

\begin{figure*}[t]
\centering
\begin{tcolorbox}[title=Feedback Example; Qwen2.5-0.5B-Instruct, promptstyle]
\lstset{
    basicstyle=\scriptsize\sffamily\scriptsize,
    breaklines=true,
    frame=none,
    columns=fullflexible,
}
\begin{lstlisting}[linewidth=\linewidth]
[Fineweb-Edu Text Truncated]

### ASSIGNMENT: ###
Consider the Bhopal disaster as a case study. What are the key social, environmental, and economic impacts of the disaster that still persist today, and how do they relate to each other?

Identify and explain the short-term and long-term health effects on humans, especially the survivors and their descendants
Discuss the economic consequences for the victims and their families, including the lack of adequate compensation and rehabilitation
Analyze the environmental degradation caused by the disaster, particularly the contamination of groundwater, and its ongoing effects on the ecosystem and human health

### STUDENT ANSWER: ###
The key social, environmental, and economic impacts of the Bhopal disaster that still persist today are closely interconnected. The disaster had an immediate impact on human health, with thousands of people killed or injured, and long-term effects including cancers, mental and physical retardation, and sterility. These health effects have had significant social consequences, including the stigmatization of survivors and their families, and economic consequences, such as the loss of livelihoods and inadequate compensation. [Truncated..]

### Feedback SEFL-tuned model ###
Overall, your analysis effectively highlights the interconnectedness of the social, environmental, and economic impacts of the Bhopal disaster. However, there are areas for improvement.
 
To strengthen your argument, consider providing specific data and statistics to illustrate the long-term health effects of the Bhopal disaster. For example, you could mention the increased risk of cancer, neurological damage, and reproductive problems in the survivors.
In addition to discussing the long-term health effects, explore the role of environmental pollution in exacerbating existing social and economic inequalities. How does the contaminated water supply affect the daily lives of the survivors and their families? What measures can be taken to address these disparities?
To take your analysis to the next level, consider proposing concrete solutions to address the environmental degradation caused by the disaster. This could involve promoting sustainable development practices, providing access to clean water and sanitation, or establishing community-led initiatives to rebuild and rehabilitate the affected areas.

### Feedback non-tuned model ###
 **Short-Term Health Effects:**
 - **Survivors and Their Descendants:** Survivors often experience chronic illnesses such as cancer, respiratory problems, and neurological disorders. The long-term psychological impacts, including depression, anxiety, and post-traumatic stress disorder (PTSD), continue to affect survivors and their families.
 - **Public Awareness and Education:** The disaster sparked widespread public awareness about the risks associated with industrial accidents, contributing to increased scrutiny and regulation of hazardous industries. However, some survivors continue to face discrimination and stigma due to the traumatic experiences.
 
 [Truncated]
 
 **Conclusion:**
 The Bhopal disaster has left a lasting impact on society, economy, and the environment. It highlights the need for robust regulatory frameworks, technological advancements, and sustained attention to environmental protection to ensure that the lessons learned from this tragedy are applied effectively to prevent future disasters. The ongoing challenges include addressing the environmental legacy, improving access to clean water and air, and ensuring equitable distribution of resources to combat the socio-economic disparities that persist.
 ---
 This response provides a comprehensive analysis of the social, environmental, and economic impacts of the Bhopal disaster, drawing on the key points discussed in the assignment and offering insights into the broader context of environmental justice and sustainable development.

\end{lstlisting}
\end{tcolorbox}
    \caption{\textbf{Feedback Example.} We show a feedback example of the \sefl{}-tuned and non-tuned \qwenmini{} model.}
    \label{fig:feedback-example}
\end{figure*}
\end{document}